\documentclass[10pt,twocolumn,letterpaper]{article}

\usepackage{iccv}
\usepackage{times}
\usepackage{epsfig}
\usepackage{graphicx}
\usepackage{amsmath}
\usepackage{amssymb}

\usepackage{tabularx}
\usepackage{amsmath}
\usepackage{amssymb}
\usepackage{booktabs}

\usepackage{multirow}
\usepackage{adjustbox}
\usepackage{float}
\usepackage{pgfplots}
\usepackage{xr}

\usepackage[pagebackref,breaklinks,colorlinks]{hyperref}
\usepackage[hypcap=true]{subcaption}
\usepackage{comment}

\usepackage[accsupp]{axessibility}

\iccvfinalcopy % *** Uncomment this line for the final submission

\def\iccvPaperID{4715} % *** Enter the ICCV Paper ID here

\ificcvfinal\fi

\begin{document}

%%%%%%%%% TITLE
\title{Multi-event Video-Text Retrieval}

\def\institutey{LMU Munich}
\def\institutew{Munich Center for Machine Learning}
\def\institutes{Oxford University}
\author{
\textbf{Gengyuan Zhang \textsuperscript{1,2} \quad Jisen Ren \textsuperscript{1} \quad Jindong Gu \textsuperscript{3}\thanks{Corresponding author} \quad Volker Tresp \textsuperscript{1,2}}  \\
\textsuperscript{1} LMU Munich, Munich, Germany\\
\textsuperscript{2} Munich Center for Machine Learning, Munich, Germany \\
\textsuperscript{3} University of Oxford, Oxford, United Kingdom \\
\tt\small zhang@dbs.ifi.lmu.de \quad jindong.gu@outlook.com
}

\maketitle
\ificcvfinal\thispagestyle{empty}\fi

%%%%%%%%% ABSTRACT
\begin{abstract}  
Video-Text Retrieval (VTR) is a crucial multi-modal task in an era of massive video-text data on the Internet.
A plethora of work characterized by using a two-stream Vision-Language model architecture that learns a joint representation of video-text pairs has become a prominent approach for the VTR task.
However, these models operate under the assumption of bijective video-text correspondences and neglect a more practical scenario where video content usually encompasses multiple events, while texts like user queries or webpage metadata tend to be specific and correspond to single events.
This establishes a gap between the previous training objective and real-world applications, leading to the potential performance degradation of earlier models during inference.
In this study, we introduce the Multi-event Video-Text Retrieval (MeVTR) task, addressing scenarios in which each video contains multiple different events, as a niche scenario of the conventional Video-Text Retrieval Task. We present a simple model, Me-Retriever, which incorporates key event video representation and a new MeVTR loss for the MeVTR task. Comprehensive experiments show that this straightforward framework outperforms other models in the Video-to-Text and Text-to-Video tasks, effectively establishing a robust baseline for the MeVTR task. We believe this work serves as a strong foundation for future studies.
Code is available at \url{https://github.com/gengyuanmax/MeVTR}.

\end{abstract}

\section{Introduction}
% Video retrieval and captions retrieval
% C
With the proliferation of multimedia data on the Internet every day, Video-Text Retrieval (VTR) task gains increasing importance in searching for desired items from the opposite modality given a video or text query.
Over the past years, numerous efforts ~\cite{xu2015jointly, otani2016learning, dong2018predicting, gabeur2020multi, miech2020end, bain2021frozen, cheng2021improving, luo2022clip4clip, gao2021clip2tv, fang2021clip2video,yang2021taco, gu2023systematic} have been undertaken to improve the multi-modal retrieval performance on both Video-to-Text and Text-to-Video retrieval tasks.
Along with the popularity of powerful vision foundation models like CLIP~\cite{radford2021learning} and ALIGN~\cite{li2022align} that align the image-text pairs in a joint feature space, two-stream video-text models~\cite{luo2022clip4clip, gao2021clip2tv} targeting to learn video-text correspondences have further become a mainstream method for the Video-Text Retrieval task.

\begin{figure*}
% \vspace{-1cm}
    \includegraphics[width=\textwidth,trim={1.2cm 8cm 1.2cm 0.2cm}, scale=2]{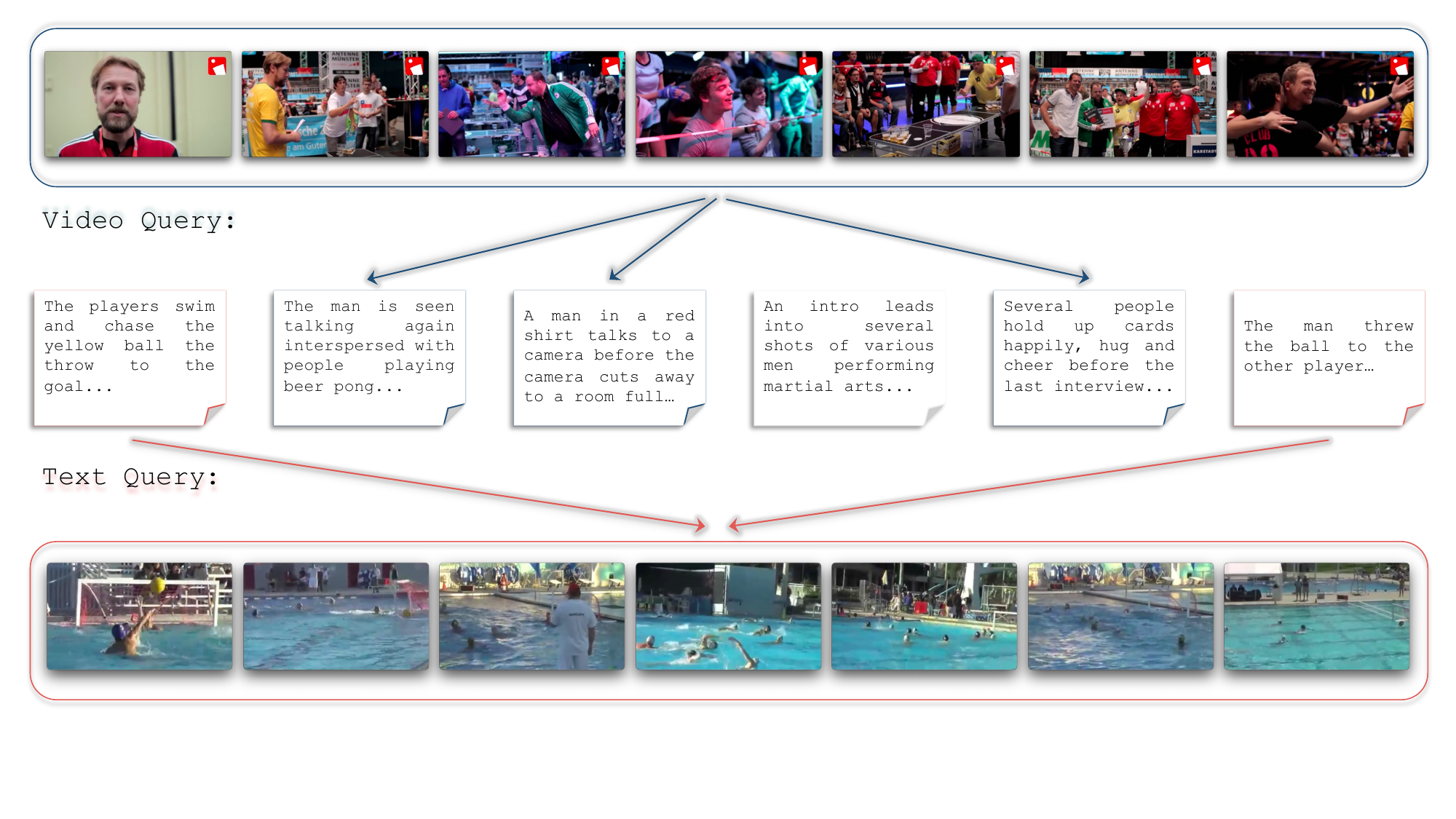}
\captionof{figure}{An example case of multi-event videos from ActivityNet~\cite{caba2015activitynet}.
The video depicts a sequence of unrelated and discontinuous events, including the progression  ``\textit{a girl is sitting on the beach}" $\rightarrow$ ``\textit{a young man is practicing tightrope walking}" $\rightarrow$ ``\textit{a scene of sunset by the beach}." Each textual caption only corresponds to a fragment of the video. Such short and specific textual captions are prevalent in our everyday video data and constitute a common video-text retrieval scenario.}
\label{fig:teaser}
\end{figure*}

However, this paradigm neglects that videos usually contain more than a single event, while a single textual caption can only capture a fragment of the entire video content. 
As depicted in Fig.~\ref{fig:teaser}, the example video spans a sequence of unrelated and discontinuous events, and each single textual caption merely corresponds to a video segment.
This congruity undoubtedly contradicts the fundamental of the traditional VTR task, which aims to establish a mapping between video-text pair.

This fact presents a prevalent scenario in Video-Text Retrieval: videos are intricate, often containing multiple events, whereas texts (typically like search queries or webpage metadata) tend to be specific and fragmentary. 
Consequently, this dichotomy leads to divergent semantic meanings for a given video-text pair. Such divergence contradicts the conventional training objective of aligning representations for video-text pairs in VTR models, that video features are aligned with the corresponding text feature.

% formulation 
In this study, we formally introduce a new and realistic VTR setting involving multi-event videos, dubbed as \textit{\textbf{M}ulti-\textbf{e}vent \textbf{V}ideo-\textbf{T}ext \textbf{R}etrieval} (\textbf{MeVTR}).
In the introduced MeVTR task, a text describes a single event within a video item, whereas a video corresponds to multiple relevant textual captions, each detailing different events.

We assess the performance of previous VTR models trained under standard VTR settings that learn a bijective video-text correspondence on the MeVTR task without retraining.
Our evaluation reveals distinct levels of performance degradation, as illustrated in Table~\ref{tab:compare}. This observation confirms the presence of a performance gap between training within the conventional VTR framework and real-world MeVTR inference scenarios.

% challenge
Besides, we retrain a series of previous models to adapt to the MeVTR task. However, we find that they cannot perform equally well on the Video-to-Text and Text-to-Video.
We speculate that this is caused by non-injective video-text correspondences in the MeVTR setting, where previous VTR models embed a video and multiple distinct caption texts into a joint representation space and aim at aligning the video feature with the features of all caption texts. Consequently, distinct texts corresponding to the same video are mapped to the same feature, potentially resulting in suboptimal retrieval performance.

Under this circumstance, we introduce a new CLIP-based model dubbed Me-Retriever tailored for the MeVTR task. Considering the dichotomy of multi-event video-text pairs, we formulate our approach as follows: (1) representing videos by a bag of key event features and (2) employing a multi-event video-text retrieval loss. This strategy serves the dual purpose of mitigating the collapse of textual features and harmonizing the learning objectives between the Text-to-Video and Video-to-Text tasks.

% contributions
Our contributions can be summarized as follows:
\begin{enumerate}
    \item We introduce a new task, Multi-event Video-Text Retrieval for retrieving multi-event video-text pairs, and define new evaluation metrics in MeVTR;
    \item We propose a new model called Me-Retriever that represents videos with selected key events and a MeVTR loss for training on MeVTR;
    \item We conduct comprehensive experiments to showcase Me-Retriever's effectiveness and to establish it as a simple yet robust baseline for future research.
\end{enumerate}

\begin{table*}
\centering
    \resizebox{\textwidth}{!}{
    \setlength\heavyrulewidth{0.25ex}
\setlength\lightrulewidth{0.2ex}
    \begin{tabular}{lcccccccc}
    \toprule 
    \multicolumn{0}{l}{} & \multicolumn{4}{c}{\textbf{Video-to-Text}} & \multicolumn{4}{c}{\textbf{Text-to-Video}} \\
    \midrule
    \multicolumn{1}{l}{Model} & \multicolumn{1}{c}{Median Rank $\downarrow$} & \multicolumn{1}{c}{k$=$1 $\uparrow$} & \multicolumn{1}{c}{k$=$5 $\uparrow$} & \multicolumn{1}{c}{k$=$10 $\uparrow$} & \multicolumn{1}{c}{Median Rank $\downarrow$} & \multicolumn{1}{c}{k$=$1 $\uparrow$} & \multicolumn{1}{c}{k$=$5 $\uparrow$} & \multicolumn{1}{c}{k$=$10 $\uparrow$}  \\
    \midrule
    FROZEN-in-time & 7.0 $\nearrow$ 109.5&	15.3 $\searrow$ 0.8 & 43.3 $\searrow$ 3.1  	& 43.3 $\searrow$  5.6 & 7.0 $\nearrow$ 111.0 & 15.9 $\searrow$  1.8 & 43.0 $\searrow$ 6.8 & 59.01 $\searrow$ 11.7    \\
    CLIPBERT & 4.0 $\nearrow$ 449.0 & 22.1 $\searrow$ 1.4  & 58.2 $\searrow$ 4.7  & 69.6 $\searrow$ 7.6 & 5.0 $\nearrow$ 153.0 &	25.9 $\searrow$ 2.6	&  55.4 $\searrow$  9.5 &	69.8 $\searrow$ 15.1 \\  
    CLIP4Clip & 2.0 $\nearrow$ 53.0  &  42.5 $\searrow$ 7.5 & 74.1 $\searrow$ 19.8  & 85.8 $\searrow$ 28.4  & 13.0 $\nearrow$ 53.0 & 14.8 $\searrow$ 7.5  & 34.2  $\searrow$ 19.8 & 45.6 $\searrow$ 28.4\\
    CenterCLIP & 2.0 $\nearrow$ 51.0 & 42.8 $\searrow$ 7.8 & 73.8 $\searrow$ 20.8 & 85.3 $\searrow$ 29.2 & 2.0 $\nearrow$  13.0 &41.8 $\searrow$	15.1 &	73.9 $\searrow$ 34.6 & 84.7 $\searrow$ 46.4\\
    % Singularity & - $\searrow$ 84.0 & - $\searrow$	5.31 	& - $\searrow$ 15.09 & - $\searrow$ 22.01 & - $\searrow$ 25.0	&  30.8 $\searrow$ 9.03 &	55.9 $\searrow$ 23.95 & 66.3 $\searrow$ 	34.33\\ 
    % Ours(attn) & 45.5 & 5.27 / 15.88 / - & 17.06 / 41.47 / 2.23 & 26.47 / 57.43 / 5.43 & 55.91 / 88.47 / 23.45 \\
 %& 47.0 & 7.12/23.39/- & 19.40/48.04/2.38  & 27.69/61.23/4.96 & 55.64/89.75/22.29  \\
    \bottomrule
\end{tabular}}
\caption{Comparison of model performance on VTR and MeVTR tasks on ActivityNet Captions. We train each model on the ActivityNet Captions with its original codes and evaluate them on MeVTR. We show the performance of VTR on the lefthand of $\searrow$($\nearrow$) and the performance of MeVTR on the right. $\uparrow$ means higher the better, and $\downarrow$ means lower the better. It is shown that model performance deteriorates to a large extent on both Video-to-Text and Text-to-Video tasks.} 
\label{tab:compare}
\end{table*}

\section{Related Work}
\vspace{0.1cm}
\noindent\textbf{Video Language Model}
%Vision-Language model
A series of work on Vision-Language Models (VLM) has been made like~\cite{li2019visualbert,lu2019vilbert,chen2020uniter,zhou2020unified,zhang2021vinvl} for multi-modal representation learning.
%video language model
Given that videos are a structured sequence of images and typically entail higher computational demands than other visual data formats, the domain of Video-Language models has garnered significant attention. 
Notable contributions within this category include works by~\cite{sun2019videobert,li2020unicoder,su2019vl, xu2015jointly, lu2019vilbert, li2022align, tsimpoukelli2021multimodal, ju2022prompting, fu2021violet} which aim at video foundation models for video tasks. These models represent a subcategory of VLMs, adept at processing video and text inputs.

\vspace{0.2cm}
\noindent\textbf{Video Representation}
% anothre taxonomy wrt ways of processing videos
In a line of work~\cite{tran2015learning, ying2020deformable, huang2020efficient, li20212d}, convolutions are performed on videos to fuse spatial and temporal information. 
Meanwhile, as transformers~\cite{vaswani2017attention} have achieved success firstly in text~\cite{devlin2018bert}, and then vision fields ~\cite{dosovitskiy2020image, liu2021swin, touvron2021training, chen2021visformer}, attempts to introduce transformers to extract video representations have been made in~\cite{sun2019learning, arnab2021vivit, wu2021cvt, zhou2021deepvit, neimark2021video,  liu2022video}.
% pre-trained-model
The success of pre-trained Vision-Language models has significantly impacted video representation learning. Vision-Language models ~\cite{lu2019vilbert, su2019vl, miech2020end,radford2021learning} have demonstrated that textual supervision is conducive to visual representation and to a uniform video-language representation. Among them, the success of representation models~\cite{lei2021less, gao2021clip2tv, wang2023all, wang2022internvideo} has bridged the gap between video and text representations.

\vspace{0.2cm}
\noindent\textbf{Video-Text Retrieval}
Video-Text Retrieval is a fundamental task of multi-modal learning, posing a good proxy task for learning joint representations of videos and texts.
% Previous work on video-text retrieval designed intensive fusion mechanisms for cross-modal learning. Recently, pre-trained
% models. A line of work ~\cite{} has dominated the leaderboard of video-text retrieval. with noticeable results on zero-shot retrieval.
% Existing research on video-text retrieval primarily follows two distinct approaches: models incorporating Mixture of Expert (MoE) structures, as evident in works like~\cite{xu2015jointly, miech2018learning, gabeur2020multi, cheng2021improving, dzabraev2021mdmmt}, strive to leverage all accessible resources, encompassing visual, textual, audio, and other sensory data. 
Two-stream Video-Text models~\cite{dong2021dual, dong2018predicting, yang2021taco, patrick2020support} concentrate on processing video-text pairs as inputs and aligning visual and textual representations within a joint representation space.
For instance, Frozen~\cite{bain2021frozen} employs a visual encoder that can be adaptively trained on images and videos, incorporating temporal contexts via curriculum learning. TACo~\cite{yang2021taco}, in contrast, focuses on aligning content words in texts with corresponding visual content, adding an extra learning objective.
With the surge of large Vision-Language foundation models representative of CLIP~\cite{radford2021learning}, models~\cite{luo2022clip4clip, gao2021clip2tv, fang2021clip2video, portillo2021straightforward, bain2021frozen, xue2022clip}, leverage the potent pre-trained model CLIP and endeavor to extend its capabilities from static images to the domain of videos.

\section{Problem}
\subsection{Problem Formulation}
Given a video-text dataset with multi-event videos $\mathcal{V}$ and texts $\mathcal{T}$, where a video $v_i\in \mathcal{V}$ is described by a set of different texts $\mathcal{T}_i\subset \mathcal{T}$ and a text is denoted as $t_j$, our task consists in retrieving a video given a text query $t^q$ and retrieving all relevant texts given a video query $v^q$. A generic Video-Text Retrieval model is to learn a similarity function $sim(v_i, t_j)$ that scores the similarity between any relevant video-text pair higher and ranks all the candidate items.

\subsection{Evaluation Metrics}
Video-Text retrieval is a recall-focused retrieval task.  A single item is retrieved in traditional settings, and the average Recall at the top rank $k$ is calculated. 
Nevertheless, in the MeVTR setting, we also want to evaluate retrieving multiple different text items given one video.
Hence we devise the following metrics: 
(1) $\textbf{Recall@k-Average}$ as the fraction of the number of correspondences ranked at top $k$;
(2) $\textbf{Recall@k-One-Hit}$ as an indication of whether any of the correspondences will be ranked in the top $k$; 
(3) $\textbf{Recall@k-All-Hit}$ as an indication of whether all items can be retrieved within the top $k$ rank. All metrics are averaged on the whole test set.
In the follow-up experiments, we report the Recall for the Text-to-Video tasks and the metrics mentioned above for the Video-to-Text tasks.

% \subsection{Analysis}
\subsection{Textual Feature Collapse}
To investigate the textual feature collapse in previous models to elucidate the performance drop by calculating the textual feature's similarity after retraining, we define the average text similarity of a video $v_i$ with $n_{v_i}$ captions as:
\begin{equation}
    sim_{t}(v_{i}) = \sum_{m=1}^{N_{v_i}} \sum_{n=1}^{N_{v_i}}  \frac{1}{N_{v_i}^2} sim(t_m, t_n) 
\end{equation}

\begin{figure}
\centering
% \resizebox{\textwidth}{}{
% \ctikzfig{st}
% }
\includegraphics[scale=0.5]{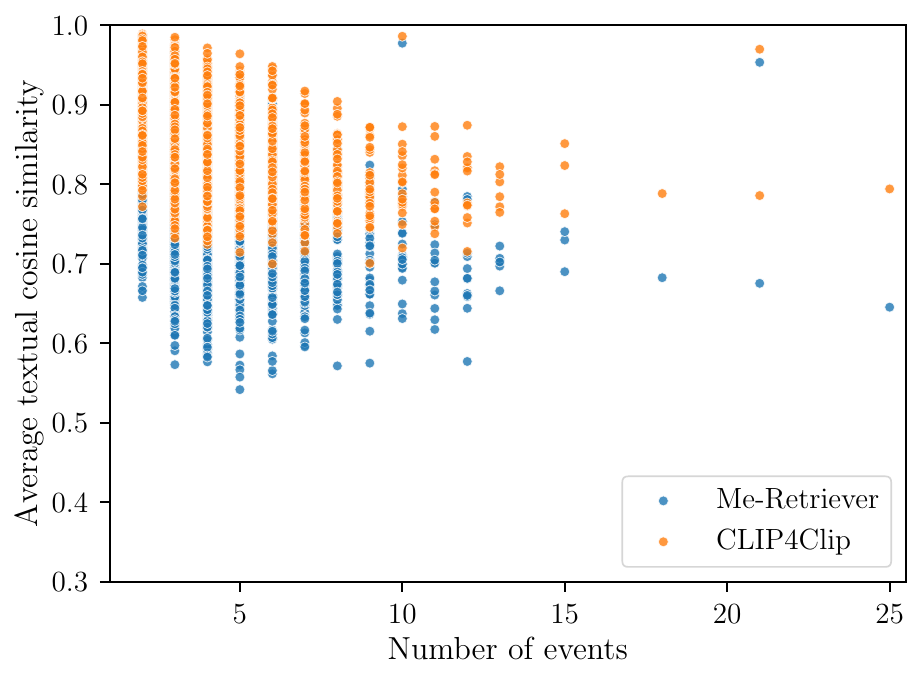}
\caption{We compare the average cosine similarity between all text pairs of videos with a different number of events. Me-Retriever can generate more diverse text features than CLIP4Clip and refrain from text features collapsing, as we discuss in the main part.}
%We use the \textbf{dynamic} weighting coefficient as a baseline and plot the performance difference of different fixed scales. Negative values indicates a performance drop compared with the dynamic weighting coefficient.}
\label{fig:compare}
\end{figure}

In Fig.~\ref{fig:compare}, we attempt to visualize the average text similarity of all videos with a different number of events. Different texts are expected to have lower sentence similarity to retain their meanings due to retraining on MeVTR.
We chose CLIP4Clip for comparison in our experiments, as it still performs relatively well in MeVTR tasks compared to other models. 
As a comparison, we show that Me-Retriever, which will be introduced in the next section, can generate more diverse textual features than CLIP4Clip and effectively avoid textual feature collapse.

\begin{figure*}
\centering
% \resizebox{\textwidth}{}{
% \ctikzfig{st}
% }
\includegraphics[width=\textwidth, trim={1cm 5cm 1cm 1cm},clip]{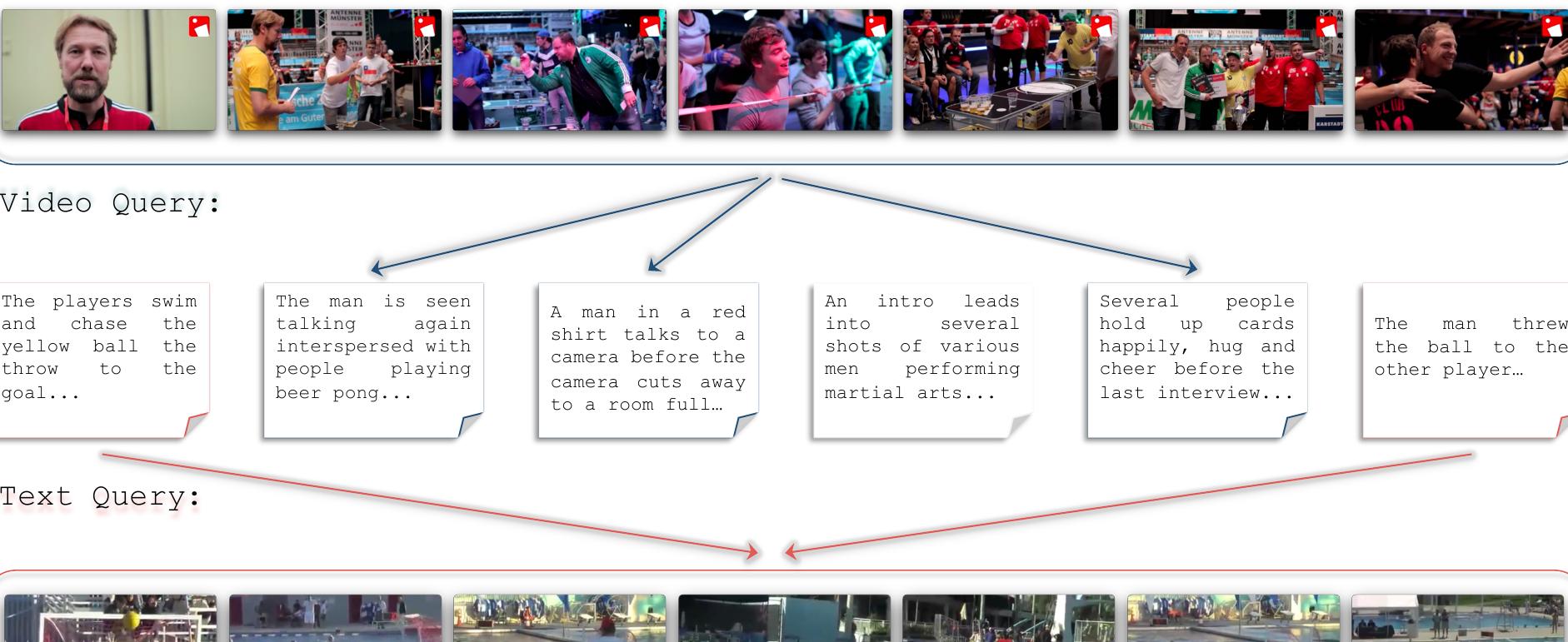}
\caption{The overall framework of Me-Retriever. The model adopts CLIP~\cite{radford2021learning}'s Visual Encoder(VE) and Text Encoder(TE). After the Visual Encoder, [\texttt{CLASS}] tokens in the last hidden layer are taken as frame embeddings. We use a clustering-based Key Event Selection module to aggregate similar frames and extract key events. Each textual caption is fed into Text Encoder, and [\texttt{EOS}] will be used as text embedding. The similarity between these key events of any video $v_i$ and any textual caption $t_j$ is measured in the Similarity Calculator. For each video, there are multiple text correspondences as positive samples.}
\label{fig:model}
\end{figure*}

\section{Method}
We propose Me-Retriever consisting of a Key Event Selection module and MeVTR loss as shown in Fig.~\ref{fig:model}.
\subsection{Key Event Video Representation}

Intuitively, human brains do not remember experiences with all continuous scenes and consolidate all the details. Instead, brains retain memories of certain discrete events ~\cite{zheng2022neurons}. This insight motivates us to represent a video with a bag of frames as key events.
For leveraging frame embeddings extracted from the CLIP visual encoder $\{\mathbf{f}^1_i, \mathbf{f}^2_i, \dots, \mathbf{f}^{|v_i|}_i\}$, we define the Key Event Selection module that selects the most distinctive keyframes.
% We attempt to put a clustering-based approach into practice to combine frames close in the feature space and take representatives of clusters as new representations of possible events.

% \paragraph{\exevent}
We employ a classic clustering method, K-Medoids~\cite{kaufman_partitioning_1990}.
Compared to other clustering methods, the representative of each cluster given by the K-Medoids algorithm is from the sample points, which corresponds to a particular frame and has \textit{explicit} interpretation. The criterion is minimizing the distance of all frames in the cluster to its medoid. $K$ frames are chosen as initial medoids; the remaining frames are then assigned to their nearest medoids, and within each cluster, new medoids are again selected based on cosine similarity as a metric function.
% When training, we adaptively update the parameters and cluster over frame embeddings. For clustering, the medoid assignment will be treated as a constant in backpropagation; and for parameter fine-tuning, .
Video $v_i$ can be represented as a sequence of key event embeddings $\mathbf{K}_i=\{\mathbf{k}^1_i, \mathbf{k}^2_i, \dots, \mathbf{k}^{K}_i\}$.
In the loss calculation, we use the pre-computed assignment of all frames as key event video representation.

\subsection{MeVTR Loss}\label{subsec:loss}
Following the dual learning paradigm of Text-to-Video and Video-to-Text on VTR tasks, we propose MeVTR loss for MeVTR learning.
As previously discussed, there is an issue of textual feature collapse during MeVTR training. To address this, we propose disentangling various positive instances within video samples by excluding non-self positive instances from the softmax function.
Moreover, considering the dissimilar granularity between multi-event videos and single-event text, we propose adopting a dynamic weighting strategy. This strategy balances the loss scale of both learning objectives during the training process.

MeVTR loss is defined as a weighted sum of $\mathcal L_{t2v}$ and $\mathcal L_{v2t}$ as follows:
\begin{equation}
    \mathcal{L}_{MeVTR} = \mathcal L_{v2t} + \alpha \mathcal L_{t2v}
\end{equation}

\begin{table*}\tiny
\centering
    \resizebox{0.98\textwidth}{!}{
\setlength\heavyrulewidth{0.25ex}
\setlength\lightrulewidth{0.2ex}
    
    \begin{tabular}{lccccc}
    \toprule 
    \multicolumn{6}{c}{\textbf{Video-to-Text}} \\
    \midrule
    \multicolumn{1}{l}{Model} & \multicolumn{1}{c}{Median Rank $\downarrow$} & \multicolumn{1}{c}{k$=$1 $\uparrow$} & \multicolumn{1}{c}{k$=$5 $\uparrow$} & \multicolumn{1}{c}{k$=$10 $\uparrow$} & \multicolumn{1}{c}{k$=$50 $\uparrow$} \\
    \midrule
    % CLIP4Clip(ZS)~\cite{luo_clip4clip_2021} & 53.0 & 7.54 / 23.67 / - & 19.89 / 48.46 / 2.50 & 28.44 / 62.38 / 5.31 & 53.56 / 88.39 / 20.48\\
    CLIP4Clip(mean) &  39.0 & 6.60 / 20.74 / - & 20.32 / 47.02 / 3.46 & 29.37 / 61.26 / 6.71 & 58.41 / 89.77 / 26.24\\
    CLIP4Clip(tight) & 49.5 & 5.23 / 16.45 / - & 16.74 / 41.94 / 1.77 & 25.50 / 57.37 / 4.23 & 54.22 / 88.00 / 21.52 \\  
    CenterCLIP & 38.0 & 6.93 / 21.62 / - &  20.58 / 47.98 / 3.66 & 29.90 / 61.72 / 6.96 & 58.88 / 90.34 / 26.28 \\
    % FROZEN-in-time(ZS) &109.5&	0.80 / 2.58 / - &	3.10 / 9.17 / 0.10	& 5.60 / 15.78 / 0.33	& 19.31 / 43.99 / 3.34 \\ 
    FROZEN-in-time & 330.0 &	0.30 / 0.67 / -	& 0.83 / 2.10 / 0.22 &	1.60 / 3.37 / 0.67	& 5.89 / 12.97 / 2.55  \\
    % CLIPBERT(ZS)& 449.0 & 1.44 / 4.51 / - & 4.72 / 13.69 / 0.18	& 7.68 / 21.19 / 0.47	& 21.51 / 47.55 / 3.99 \\
    CLIPBERT & 174.5&	1.80 / 6.10 / - &	6.48 / 18.97 / 0.18 &	10.93 / 29.00 / 0.87 & 31.34 / 61.42 / 8.72  \\ 
    % Singularity(ZS) & 84.0&	5.31 / 16.68 / -	& 15.09 / 38.84 / 1.57	& 22.01 / 51.80 / 3.21 & 45.70 / 81.47 / 15.15 \\  
    Singularity & 64.5	& 4.83 / 14.68 / - &	15.1 / 37.79 / 1.95 &	22.23 / 49.50 / 4.41 & 48.96 / 81.94 / 19.38  \\
    \midrule
    Ours(avg) & \textbf{35.0}  & \textbf{8.52} / \textbf{26.91} / - & \textbf{23.56} / \textbf{54.18} / \textbf{3.86} & \textbf{32.95} / \textbf{67.13} / \textbf{7.79} & \textbf{61.26} / \textbf{92.50} / \textbf{27.80} \\
    Ours(max) & 49.0 & 6.98 / 22.72 / - & 18.72 / 46.37 / 2.30 & 27.04 / 60.62 / 5.76 & 54.19 / 88.37 / 20.90\\
    % Ours(attn) & 45.5 & 5.27 / 15.88 / - & 17.06 / 41.47 / 2.23 & 26.47 / 57.43 / 5.43 & 55.91 / 88.47 / 23.45 \\
 %& 47.0 & 7.12/23.39/- & 19.40/48.04/2.38  & 27.69/61.23/4.96 & 55.64/89.75/22.29  \\
\bottomrule
\end{tabular}}
\caption{Results of the Video-to-Text retrieval task on ActivityNet Captions dataset. %The baseline methods are CLIP4Clip ~\cite{luo_clip4clip_2021} and CenterCLIP ~\cite{zhao_centerclip_2022}.% 
In each column, we report the $\text{Recall}@k$-Average/One-Hit/All-Hit for each $k={1,5,10,50}$ respectively. The best result in each column is emphasized in \textbf{bold}. We ignore the $\text{Recall}@1$-All-Hit since all models can only achieve a nearly zero result. Ours(avg) maintains its advantage in the text retrieval task and surpasses all other models by a large margin.} 
\label{tab:basic-results-v2t-an}
\end{table*}

\begin{table*}\tiny
\centering
    \resizebox{0.98\textwidth}{!}{
\setlength\heavyrulewidth{0.25ex}
\setlength\lightrulewidth{0.2ex}
    
    \begin{tabular}{lccccc}
    \toprule 
    \multicolumn{6}{c}{\textbf{Video-to-Text}} \\
    \midrule
    \multicolumn{1}{l}{Model} & \multicolumn{1}{c}{Median Rank $\downarrow$} & \multicolumn{1}{c}{k$=$1 $\uparrow$} & \multicolumn{1}{c}{k$=$5 $\uparrow$} & \multicolumn{1}{c}{k$=$10 $\uparrow$} & \multicolumn{1}{c}{k$=$50 $\uparrow$} \\
    \midrule

    CLIP4Clip(mean) & 441.5 & 1.32 / 2.92 / - & 4.07 / 8.32 / 1.80 & 5.65 / 11.84 / 2.55 & 16.84 / 32.01 / 7.65\\
    CLIP4Clip(tight) & 391.0 & 0.80 / 1.80 / - & 2.45 / 6.30 / 0.75 & 4.19 / 10.27 / 1.35 & 15.26 / 30.43 / 6.37  \\  
    CenterCLIP &  408.0 & 1.32 / 3.00 / - & 3.71 / 7.87 / 1.57 & 6.33 / 13.42 / 2.62 & 17.37 / 33.43 / 7.72 \\
    FROZEN-in-time  & 810.0 &	0.30 / 0.67 / - &	0.83 / 2.10 / 0.22 &	1.60 / 3.37 / 0.67	& 5.89 / 12.97 / 2.55\\
    ClipBERT & 1339.5 & 0.15 / 0.45 / - &	0.60 / 1.50 / 0.15	& 1.19 / 3.30 / 0.22 & 5.31 / 11.92 / 1.87 \\
    Singularity & 621.5 &	0.95 / 2.17 / - &	2.84 / 5.92 / 1.42	& 4.35 / 9.52 / 1.95	& 12.22 / 25.94 / 4.87 \\
    \midrule
    Ours(avg) & 406.0 & \textbf{1.39} / \textbf{3.22} / - & \textbf{5.04} / \textbf{9.90} / \textbf{2.47} &	\textbf{7.02} / \textbf{13.72} / \textbf{3.37} & 17.62 / 32.08 / \textbf{8.70}\\
    Ours(max) & \textbf{365.6} & 1.01 / 2.62 / - & 4.03 / 9.00 / 1.72	& 6.39 / 13.72 / 2.70 & \textbf{19.08} / \textbf{36.66} / 8.10\\
    % Ours(attn)  & 653.0 & 0.69 / 1.50 / - & 1.76 / 3.90 / 0.67 & 3.44 / 7.72 / 1.35 & 10.49 / 21.96 / 4.35\\
    
\bottomrule
\end{tabular}}
\caption{Results of the Video-to-Text retrieval task on Charades-Event dataset. We use the same notation as in Tab.~\ref{tab:basic-results-v2t-an}. Me-Retriever(avg) and Me-Retriever(max) can achieve best performance.}
\label{tab:basic-results-v2t-charades}
\end{table*}

% The prior normalized softmax function does not apply in a multi-event scenario since there will be a variable amount of heterogeneous texts in correspondence for each video instance. 
% % The issue is twofold, as we will discuss: (1) multiple heterogeneous positive instances will conflict with each other in a softmax function; (2) Video-to-Text loss and Text-to-Video loss are unbalanced in their scale and these two learning objectives are asymmetric.
% Therefore, we propose MeVTR Loss for training, which is formulated as:

where $\mathcal{L}_{v2t}$ and $\mathcal{L}_{t2v}$ are respectively formulated as:

\begin{equation}
    \mathcal{L}_{v2t} = -\frac{1}{|\mathcal{B}|} \sum^{|\mathcal{B}|}_{i}  \sum^{|\mathcal{T}_i|}_{k} \frac{1}{|\mathcal{T}_i|} \log \frac{\exp{sim(\mathbf{K}_i, \mathbf{w}_k)/\tau}}{\sum^{\mathcal{B}^{-}\cup \{k\}}_{j=1}\exp{sim(\mathbf{K}_i, \mathbf{w}_j)/\tau}},
\end{equation}
 
\begin{equation}
    \mathcal{L}_{t2v} = -\frac{1}{|\mathcal{B}|} \sum^{|\mathcal{B}|}_{i} \log \frac{\exp{sim(\mathbf{K}_i, \mathbf{w}_i)/\tau}}{\sum^{\mathcal{B}}_{i=1}\exp{sim(\mathbf{K}_j, \mathbf{w}_i))/\tau}},
\end{equation}

\vspace{0.5cm}

$sim(\cdot)$ is the similarity function, $\mathcal{B}$ is the set of samples in the training batch, $\mathcal{B}^{-}$ is the set of negative samples in the batch, $\tau$ is the temperature coefficient, and $\mathcal{T}_i$ is the set of positive instances of one sample video in $\mathcal{L}_{t2v}$.
Notably, given a video item $v_i$, we discard all positive instances in $\mathcal{T}_i$ other than the current training sample $t_k$. This encourages similarities between a video and its all positive instances to be increased. In $\mathcal{L}_{t2v}$, it degenerates to a normal softmax function.

%Inspired by ~\cite{luo_clip4clip_2021}, we implemented a parameter-free similarity function and transformer-based learnable similarity functional to score the video-text pairs.

We denote $\alpha$ as the weighting coefficient that balances $\mathcal{L}_{t2v}$ and $\mathcal{L}_{v2t}$. Given that the two losses exhibit differing scales during training, which owes to the positive/negative sample ratio asymmetry, determining the value of $\alpha$ requires employing certain hyperparameter-tuning techniques to ensure balanced performance on both tasks.
We suggest employing a dynamic weighting strategy as opposed to a fixed weight. A dynamic weighting coefficient $\alpha$ is computed as the ratio of $\mathcal{L}_{v2t}$ to $\mathcal{L}_{t2v}$ during training. In Section~\ref{subsec:invest-loss}, we conduct experimental studies to compare the impact of various weighting strategies.

\vspace{0.2cm}
\noindent\textbf{Similarity function} 
For video-text pairs in bijective correspondences, the similarity between video and text is measured by cosine similarity in the feature space.
In MeVTR, we use two approaches are used to measure the similarity between a video $v_i$ consisting of key events$\{\mathbf{k}^1_i, \mathbf{k}^2_i, \dots, \mathbf{k}^{K}_i\}$ and a text $t_j$: 
(1) we average the cosine similarity scores of all key event embeddings and the text as in Eq.~\ref{eq:avg};
(2) we select the maximum cosine similarity of any key event-text pair to represent the similarity between the video to which these medoids belong as in Eq.~\ref{eq:max}; 

\vspace{-4mm}
% \begin{equation}\label{eq:avg}
%     s_{avg}(v_i, t_j) = \mathtt{cos}(\mathtt{average}(\mathbf{h}^{i}_{1},\mathbf{h}^{i}_{1},\dots, \mathbf{h}^{i}_{k}), t_j)
% \end{equation}

\begin{equation}\label{eq:avg}
    s_{avg}(v_i, t_j) = \mathtt{avg}(\mathtt{cos}(\mathbf{k}^{i}_{1},\mathbf{w}_j), \dots, \mathtt{cos}(\mathbf{k}^{i}_{k}, \mathbf{w}_j))
\end{equation}

\vspace{-6mm}

\begin{equation}\label{eq:max}
    s_{max}(v_i, t_j) = \mathtt{max}(\mathtt{cos}(\mathbf{k}^{i}_{1},\mathbf{w}_j), \dots, \mathtt{cos}(\mathbf{k}^{i}_{k}, \mathbf{w}_j))
\end{equation}

% \vspace{-6mm}

% \begin{equation}\label{eq:attn}
%     s_{attn}(v_i, t_j) = \mathtt{transformer}(\mathbf{k}^{i}_{1},\mathbf{k}^{i}_{1},\dots, \mathbf{k}^{i}_{k}, \mathbf{w}_j)
% \end{equation}

We denote Me-Retriever with different similarity functions as Me-Retriever(avg) and Me-Retriever(max).

\section{Experiments}
% In this section, we train Me-Retriever on MeVTR settings.

% \begin{figure}
% \centering
% \hspace{-0.5cm}
% \subcaptionbox{ActivityNet\label{hist_a:sfig:a}}{
% \includegraphics[scale=0.24]{charts/act_train_duration.pdf}} 
% \hspace{-0.5cm}
% \subcaptionbox{ActivityNet\label{hist_a:sfig:b}}{
% \includegraphics[scale=0.24]{charts/act_train_event.pdf}
% } 

% \hspace{-0.5cm}
% \subcaptionbox{Charades-Event\label{hist_a:sfig:c}}{
% \includegraphics[scale=0.24]{charts/cha_train_duration.pdf}
% }
% \hspace{-0.5cm}
% \subcaptionbox{Charades-Event\label{hist_a:sfig:d}}{
% \includegraphics[scale=0.24]{charts/cha_train_event.pdf}
% }

% \caption{Histogram of training sets of ActivityNet and Charades-Event. Fig.~\ref{hist_a:sfig:a}-~\ref{hist_a:sfig:b}: ActivityNet histogram concerning video duration and number of events; Fig.~\ref{hist_a:sfig:c}-~\ref{hist_a:sfig:d}: Charades-Event histogram concerning video duration and number of events.}
% \label{fig:hist:a}
% \end{figure}

\subsection{Experiment Details}
\vspace{0.2cm}
\noindent\textbf{Datasets}
We perform experiments on two datasets containing multi-event videos: ActivityNet Captions and Charades-Event. It's worth noting that datasets such as MSR-VTT~\cite{xu2016msr}, even though videos are accompanied by multiple similar captions, do not align with our multi-event scenario.

\noindent\textbf{ActivityNet Captions} ~\cite{krishna2017dense} is a large-scale dataset with 20,000 video-text pairs extracted by web crawling on YouTube based on ActivityNet~\cite{caba2015activitynet} videos. It covers a wide range of daily scenarios. The training set contains 10,009 videos and 4,917 videos/17,505 texts in the publicly accessible evaluation set `val1'. Each video has multiple textual descriptions with time-interval annotation. This helps us construct a multi-event scenario for retrieval by taking all textual captions of a video as its associated events.

% \paragraph{Charades-Event}
\noindent\textbf{Charades-Event} is a video dataset of indoor activities with extensive action classes that we adopt from ~\cite{sigurdsson2016hollywood}.
To create a MeVTR dataset, we filter out all videos with text annotation provided by ~\cite{gao2017tall} and interpret each text as an \textit{event} in our MeVTR setting.
There are 5,338 videos/12,408 texts in the training set and 1,334 videos/3,720 texts in the test set.

\vspace{0.2cm}
\noindent\textbf{Experiment Settings}
In the key event selection module, we use 60 as the maximum number of iterations and $10^{-5}$ as the distance threshold for clustering.
Our model chooses 16 as a fixed number of key events and does not use a smaller or larger number of key events because shorter sequences cannot cover the number of events in part of the videos (the maximum event number in ActivityNet Captions is 27), and longer sequences contradict our intention of clustering.
For each experiment, we set the epoch number as 5.

\begin{table}
\centering
\resizebox{0.48\textwidth}{!}{
\setlength\heavyrulewidth{0.25ex}
\setlength\lightrulewidth{0.2ex}
    \begin{tabular}{lccccc}
    \toprule 
    \multicolumn{6}{c}{\textbf{Text-to-Video}}  \\
    \midrule
    \multicolumn{1}{l}{Model} & \multicolumn{1}{c}{Median Rank $\downarrow$} & \multicolumn{1}{c}{k$=$1 $\uparrow$} & \multicolumn{1}{c}{k$=$5 $\uparrow$} & \multicolumn{1}{c}{k$=$10 $\uparrow$} & \multicolumn{1}{c}{k$=$50 $\uparrow$} \\
    \midrule
        % CLIP4Clip(ZS) & 13.0 & 14.85 & 34.25 & 45.64 & 71.33\\
    CLIP4Clip(mean) & \textbf{11.0} & 15.25 & 36.43 & 49.11 & 76.64 \\
    CLIP4Clip(tight) & 15.0 & 10.14 & 29.07 & 41.93 & 72.50 \\
    % CLIP4Clip(T MIL) & \\ 
    CenterCLIP & \textbf{11.0} & \underline{15.53} & \textbf{37.21} & \textbf{49.78} & \textbf{77.19} \\
    % FROZEN-in-time(ZS) & 34.0 & 5.32	& 18.08 & 28.07 & 56.54 \\ 
    FROZEN-in-time & 111.0 & 1.82 & 6.80 & 11.76 & 33.86 \\
    % ClipBERT(ZS) & 13.0 & 2.60& 9.55 & 15.10 & 43.51\\
    % CLIP4Clip(T MIL) & \\ 
    ClipBERT & 48.0 & 4.20 & 14.57 & 23.35 & 51.02 \\
    % Singularity(ZS) & 25.0 &9.03  & 23.95 &	34.33	& 62.37 \\
    Singularity(FT) &22.0&	9.29&	24.96&	36.26&	65.55 \\
    \midrule
    Ours(avg) & \textbf{11.0} & \textbf{15.99} & \underline{37.13} & \underline{49.60} & \underline{76.66}  \\
    Ours(max) & 13.0 & 14.93 & 34.44 & 46.36 & 74.65\\
    % Ours(attn) & \underline{12.0} & 12.22	& 33.33	& 46.40 & 75.51\\
     %&  12.0 & \textbf{16.13} & 35.99 & 47.84 & 75.98  \\
\bottomrule
\end{tabular}}
\caption{Results of the Text-to-Video retrieval task on ActivityNet Captions dataset. We compare CLIP4Clip~\cite{luo2022clip4clip}, CenterCLIP~\cite{zhao2022centerclip}, FROZEN-in-time~\cite{bain2021frozen}, ClipBERT~\cite{lei2021less}, and \cite{lei2022revealing}. The best result in each column is emphasized in \textbf{bold}, and the second best result is \underline{underlined}. Our Me-Retriever(avg) achieves comparably good performance.}
\label{tab:basic-results-t2v-an}
\end{table}

\begin{table}
\centering
    \resizebox{0.48\textwidth}{!}{
    \setlength\heavyrulewidth{0.25ex}
    \setlength\lightrulewidth{0.2ex}
    \begin{tabular}{lccccc}
    \toprule 
    \multicolumn{6}{c}{\textbf{Text-to-Video}}  \\
    \midrule
    \multicolumn{1}{l}{Model} & \multicolumn{1}{c}{Median Rank $\downarrow$} & \multicolumn{1}{c}{k$=$1 $\uparrow$} & \multicolumn{1}{c}{k$=$5 $\uparrow$} & \multicolumn{1}{c}{k$=$10 $\uparrow$} & \multicolumn{1}{c}{k$=$50 $\uparrow$}  \\
    \midrule 
    CLIP4Clip(mean) & 136.0 & 1.53 & 6.48 & 10.54 & 29.38 \\
    CLIP4Clip(tight) & 127.0 & 1.45 & 5.54 & 9.35 & 30.11 \\
    % CLIP4Clip(T MIL) & \\ 
    CenterCLIP &  131.0 & \underline{1.93} & 6.61 & \underline{10.67} & \underline{30.13} \\
    FRONZEN-in-Time & 277.0 & 0.35 & 1.99 & 3.82 & 14.52 \\
    ClipBERT &442.0	&0.35	&1.77	&3.31	&11.10\\
    Singularity &  179.5	& 1.53 &	5.22	& 8.52 &	23.87  \\
    \midrule
    Ours(avg) & \underline{126.0} & 1.88 & \underline{6.69} & 10.59 & 29.97  \\
    Ours(max) & \textbf{112.0} & \textbf{2.02} & \textbf{7.61} & \textbf{12.74} & \textbf{34.49}\\
    % Ours(attn) & 214.0 & 0.54 & 3.36 & 6.26 & 21.02  \\
\bottomrule
\end{tabular}}
\caption{Results of Text-to-Video retrieval task on Charades-Event dataset. We use the same notation as Tab.~\ref{tab:basic-results-v2t-an}. Me-Retriever(max) and Me-Retriever(avg) can achieve the best performance compared to other models.}
\label{tab:basic-results-t2v-charades}
\end{table}

\subsection{Results}\label{sec:result}
The experiments show that Me-Retriever performs competitively within the context of MeVTR.
On the Video-to-Text task, as presented in Tables~\ref{tab:basic-results-v2t-an} and~\ref{tab:basic-results-v2t-charades}, Me-Retriever(avg) surpasses its counterparts. This result underscores our model's capability to address both tasks across different datasets effectively.
Additionally, we observe that CLIP4Clip also attains a commendable $\text{Recall}@1$-Average performance on the Video-to-Text task compared to other baseline models. This observation sheds light on the robust transferability of the pre-trained encoder to videos. This indicates that the primary challenge in MeVTR often resides within representation alignment.

% on text to video
Our model also demonstrates comparable results on both datasets on the Text-to-Video task, as shown in Tab.~\ref{tab:basic-results-t2v-an} and Tab.~\ref{tab:basic-results-t2v-charades}.
Me-Retriever(avg) demonstrates comparable performance to baseline models like CenterCLIP across numerous metrics on both datasets. Interestingly, we also notice that Me-Retriever(max) performs best on the Charades-Event dataset but not on the ActivityNet Captions dataset. This disparity can be attributed to overfitting, warranting further investigation.
The noticeable performance of certain baseline models aligns with our conjecture that prior models can be effectively adapted to the Text-to-Video retrieval task. However, this adaptation might come at the expense of suboptimal results in the Video-to-Text task, as evident in Table~\ref{tab:basic-results-v2t-an}. 
This phenomenon can be attributed to the embeddings of different texts collapsing to become more proximate, leading to an unintentional competition among positive instances that ultimately dampens their significance.

\vspace{0.2cm}
\noindent\textbf{Performance on videos with different subsets}
In order to reveal the model performance on videos of different properties like video duration and the number of events, we partition the complete ActivityNet Caption test set into two groups of subsets. These subsets are partitioned based on the video duration and the count of events within the videos, respectively: \\
\vspace{-0.6cm}
\begin{enumerate}
    \item Video duration: test-\textbf{S} (\textbf{S}hort) with videos shorter than 1 minute, test-\textbf{M} (\textbf{M}edium) with videos between 1-2 minutes, test-\textbf{L} (\textbf{L}ong video) with videos between 2-3 minutes, and test-\textbf{XL} (e\textbf{X}tra \textbf{L}ong video) with even longer videos than 3 minutes;
    \item Number of events: test-\textbf{E1} with videos containing fewer than $4$ events, test-\textbf{E2} with 5-12 events, and test-\textbf{E3} with more than $12$ events.
\end{enumerate}

\begin{table}
\centering
    \resizebox{0.48\textwidth}{!}{
\setlength\heavyrulewidth{0.25ex}
\setlength\lightrulewidth{0.2ex}
    \begin{tabular}{lccccc}
    \toprule
    \multicolumn{6}{c}{\textbf{Text to Video}}  \\
    \midrule
    \multicolumn{1}{l}{Model} & \multicolumn{1}{c}{Median Rank $\downarrow$} & \multicolumn{1}{c}{k$=$1 $\uparrow$} & \multicolumn{1}{c}{k$=$5 $\uparrow$} & \multicolumn{1}{c}{k$=$10 $\uparrow$} & \multicolumn{1}{c}{k$=$50 $\uparrow$} \\
    \midrule
    Ours (Best) & \textbf{11.0} & \textbf{15.17} & \textbf{35.94} & \textbf{48.39} & \textbf{75.85} \\
    w/o Key Event &  12.0 & 15.03 & 35.67 & 47.94 & 75.06\\
    w/o MeVTR loss & 12.0 & 14.45 & 34.90 & 47.25 & 75.24  \\
\bottomrule
\end{tabular}}
\caption{Ablation studies on key modules of Me-Retriever training on the Text-to-Video task for ActivityNet Captions. Only the $\text{Recall}@k$-Average is reported in the main part. We find that model performance drops with removing any modules, proving its efficacy.}
\label{tab:keymoodule-t2v}
\end{table}

\begin{figure}
% \begin{center}
\centering
\subcaptionbox{On subsets of different duration\label{sfig:ab}}{
\includegraphics[scale=0.39]{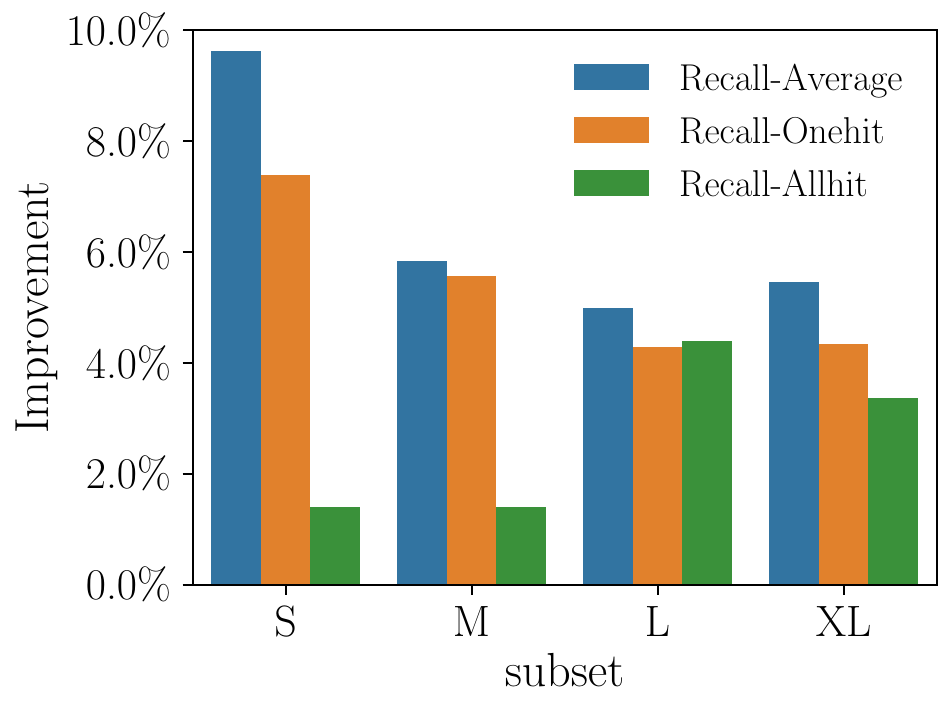}
 }
% \hspace{-0.41cm}
\vspace{0.5cm}

% \hspace{-0.49cm}
\subcaptionbox{On subsets of different event number\label{sfig:ba}}{
\includegraphics[scale=0.39]{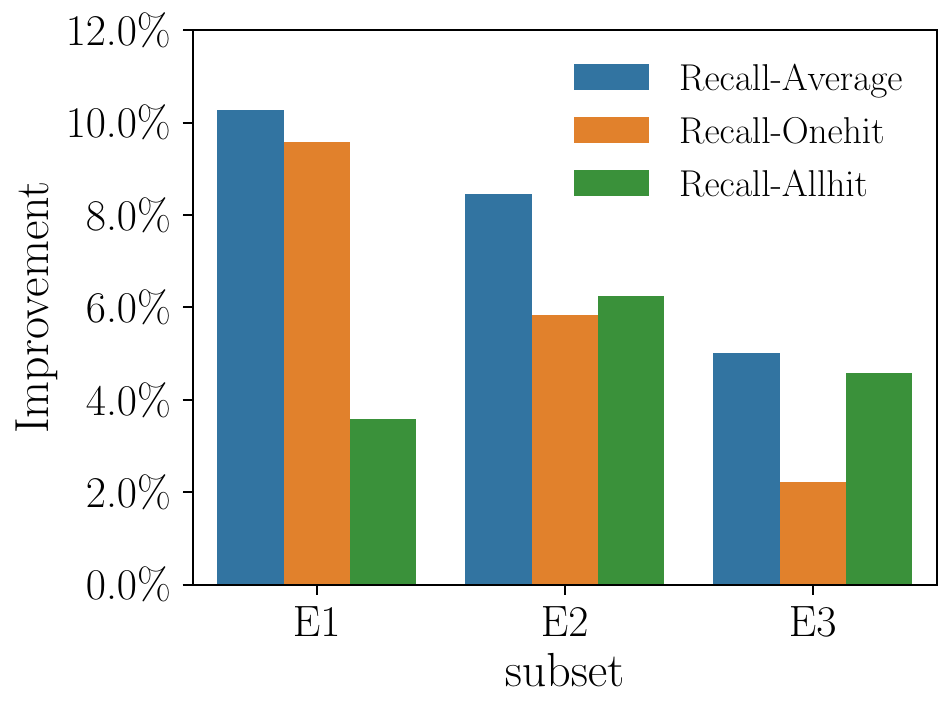}}
\caption{We compare average performance improvement in percentage($\%$) for Video-to-Text task on subsets of ActivityNet Captions. It shows how much percentage Me-Retriever(avg) is better than CLIP4Clip(mean) on different subsets.
 Fig.~\ref{sfig:ab}: Video-to-Text results for test-\textbf{S}/\textbf{M}/\textbf{L}/\textbf{XL};
 Fig.~\ref{sfig:ba}: Video-to-Text results for test-\textbf{E1}/\textbf{E2}/\textbf{E3}. }
\label{fig:compare_on_length}
% \end{center}
\end{figure}

\begin{table}
\centering
    \resizebox{0.48\textwidth}{!}{
    \setlength\heavyrulewidth{0.25ex}
\setlength\lightrulewidth{0.2ex}
    \begin{tabular}{lccccc}
    \toprule
    \multicolumn{6}{c}{\textbf{Video to Text}}  \\
    \midrule
    \multicolumn{1}{l}{Model} & \multicolumn{1}{c}{Median Rank $\downarrow$} & \multicolumn{1}{c}{k$=$1 $\uparrow$} & \multicolumn{1}{c}{k$=$5 $\uparrow$} & \multicolumn{1}{c}{k$=$10 $\uparrow$} & \multicolumn{1}{c}{k$=$50 $\uparrow$} \\
    \midrule
    Ours (Best) & \textbf{35.0} & \textbf{8.52} & \textbf{23.56} & \textbf{32.79} & \textbf{61.26} \\
    w/o Key Event &  41.5 & 7.40 & 21.00 & 30.27 & 57.71\\
    w/o MeVTR loss & 41.0 & 6.70 & 20.05 & 28.96 & 57.45 \\
\bottomrule
\end{tabular}}
\caption{Ablation studies on key modules of Me-Retriever training on the Video-to-Text task for ActivityNet Captions. Model performance declines to different extents without our modules. We conclude that our model can benefit Video-to-Text retrieval prominently.}
\label{tab:keymodule-v2t}
\end{table}

\begin{figure*}[h]
\vspace{-0.4cm}
% \begin{center}
\centering
\hspace{-0.5cm}
\subcaptionbox{Average\label{sfig1:a}}{
\includegraphics[scale=0.37]{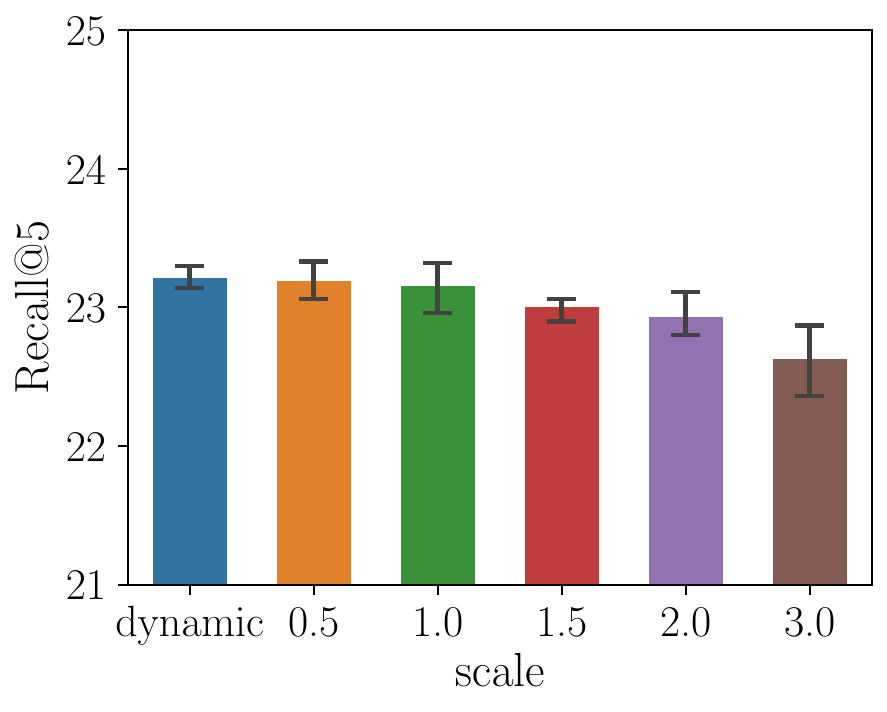}
}
\hspace{-0.1cm}
\subcaptionbox{One-Hit\label{sfig1:b}}{
\includegraphics[scale=0.37]{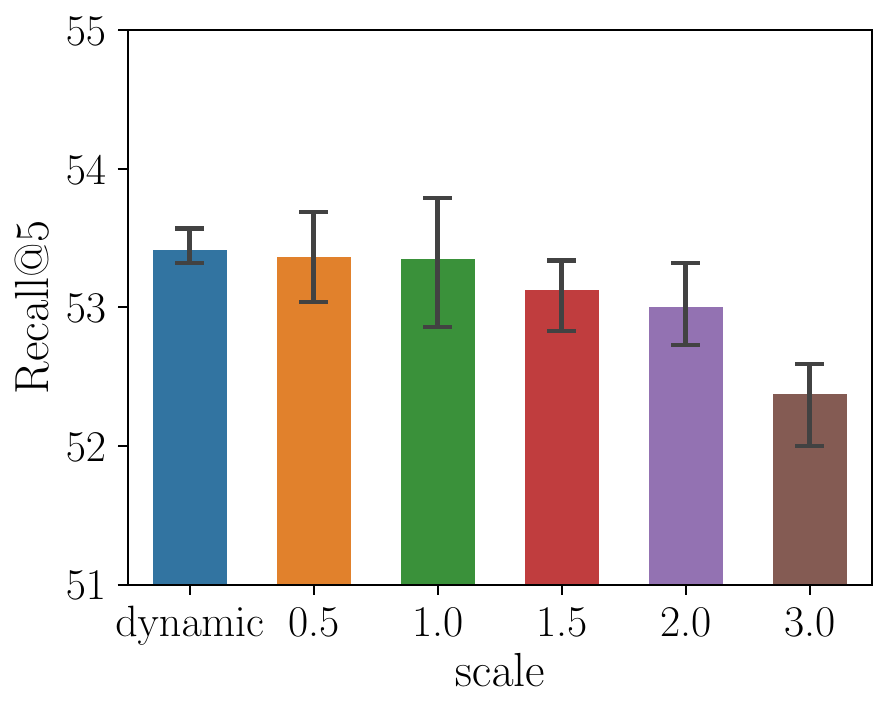}
}
\hspace{-0.1cm}
\subcaptionbox{All-Hit\label{sfig1:c}}{
\includegraphics[scale=0.37]{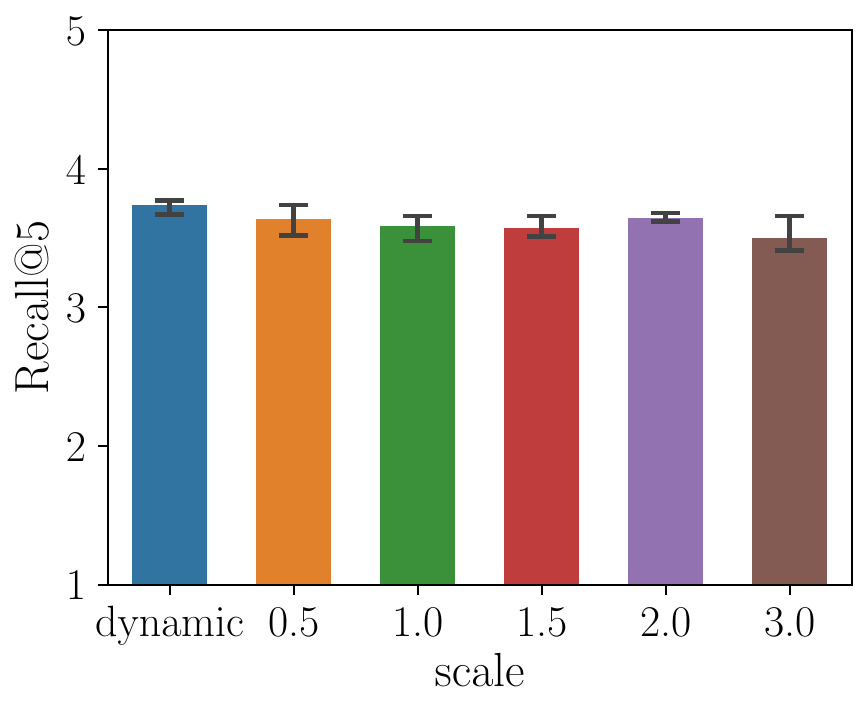}
}
\caption{We compare the model performance with different weighting strategies, a dynamic weight $\alpha$ and different choices of fixed weights from a choice of $\{0.5, 1.0, 1.5, 2.0, 3.0\}$, in the MeVTR loss on $\text{Recall}@5$ on the Video-to-Text task for ActivityNet Captions. Fig.~\ref{sfig1:a}-\ref{sfig1:c} shows the results of $\text{Recall}@5$-Average/One-Hit/All-Hit respectively. We can find that compared to a fixed weighting coefficient, a dynamic weight $\alpha$ guarantees more stable and good results in different metrics.}
\label{fig:different-scale}
% \end{center}
\end{figure*}

\begin{figure}
\centering
% \resizebox{\textwidth}{}{
% \ctikzfig{st}
% }
\includegraphics[scale=0.4]{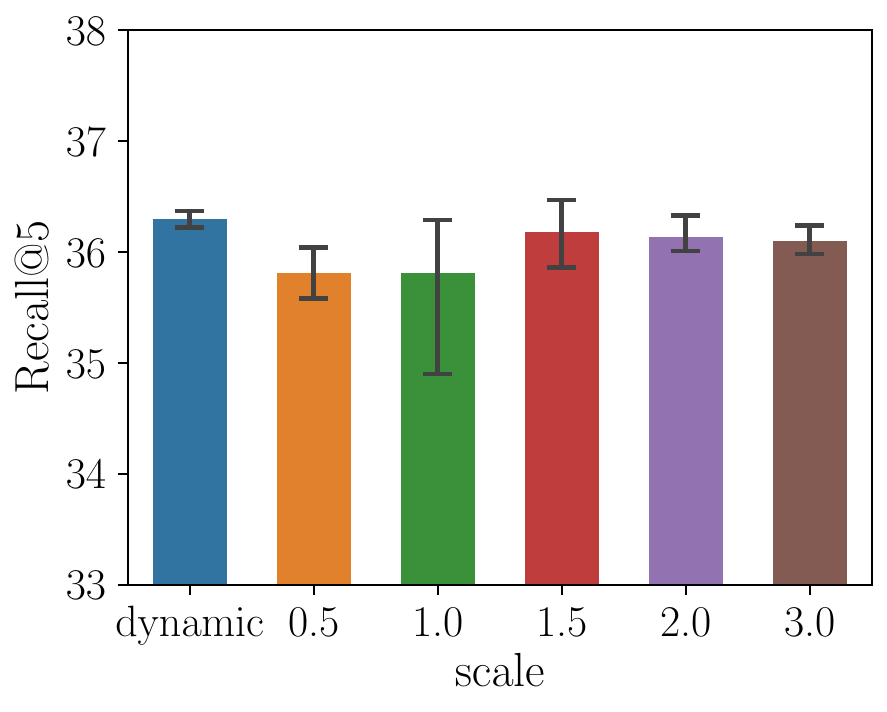}
\caption{We compare the model performance with different weighting coefficients in the MeVTR loss on $\text{Recall}@5$ on the Text-to-Video task for ActivityNet Captions. The same graphical representations are used as Fig.~\ref{fig:different-scale}. The dynamic weighting coefficient performs better than any other fixed coefficient considering performance and variance.}
%We use the \textbf{dynamic} weighting coefficient as a baseline and plot the performance difference of different fixed scales. Negative values indicates a performance drop compared with the dynamic weighting coefficient.}
\label{fig:different-scale-tvr}
\end{figure}

Comparing performance across datasets of different sizes isn't equitable for a retrieval task, as larger candidate sets exacerbate retrieval difficulties. Therefore, we gauge the average improvement in the model performance of Me-Retriever(avg) to the baseline CLIP4Clip(mean) and identify the subsets in which our model showcases notable enhancements.
From Fig.~\ref{fig:compare_on_length}, we also see that our model usually performs better on test-\textbf{E1} and test-\textbf{E2}, and improvement slightly decreases for even longer videos. We assume this problem is inevitable since we use a fixed number of clusters and videos containing many events in test-\textbf{E3}, and larger numbers of events increase the retrieval difficulties. Whereas for videos of different duration, model performance slightly varies.

\subsection{Ablation Studies}
In this section, we conduct various experiments to investigate the effectiveness of different weighting strategies in MeVTR loss and ablate the proposed components of our architecture.

% \paragraph{Impact of Finetuning CLIP layers}
\vspace{0.2cm}
\noindent\textbf{Impact of weighting strategy in MeVTR Loss}\label{subsec:invest-loss}
To study the effect of different weighting strategies of MeVTR loss, i.e., dynamic weighting and fixed weighting, we test our model trained with a set of different weighting coefficients $\alpha$ as illustrated in Fig.~\ref{fig:different-scale} and Fig.~\ref{fig:different-scale-tvr}.
On the Video-to-Text task and Text-to-Video Retrieval tasks, we notice that dynamic weighting $\alpha$ can guarantee us a comparable performance in general and less sensitivity than any fixed coefficient. It is also illustrated that when we choose a smaller coefficient, the Video-to-Text retrieval is favored, while a bigger coefficient benefits the other way around. A cutoff point is empirically between 1.0 and 2.0. The reason is that $\mathcal{L}_{v2t}$ loss always has a comparably larger scale than $\mathcal{L}_{t2v}$ loss and dominates training due to the unbalanced number of video-text instances, as we have discussed in Section ~\ref{subsec:loss}. This ensures us that a dynamic weighting strategy can improve the Text-to-Video task by amplifying the influence of $\mathcal{L}_{t2v}$ and guarantee stable improvements on both tasks.

\vspace{0.2cm}
\noindent\textbf{Effects of different modules}
We choose our best model Me-Retriever(avg) and ablate the key event selection module by removing the key event selection module and the MeVTR loss by replacing it with the cross-entropy loss, respectively. 
Results are shown in Tab.~\ref{tab:keymoodule-t2v} and Tab.~\ref{tab:keymodule-v2t}. We display the $\text{Recall}@k$-Average on both tasks in the table and observe that the model performances drop after removing any modules. 
In particular, our method demonstrates a larger improvement by $1.12\%$ without the key event selection module and $1.82\%$ without the MeVTR loss on the Video-to-Text task. This agrees with our assumption that our Me-Retriever model can deal with textual feature collapse in the MeVTR training. On the Text-to-Video task, both modules also show benefits to the performance on all metrics.

\vspace{0.2cm}
\noindent\textbf{Textual features of multi-event videos}
To show whether Me-Retriever can prevent heterogeneous text features of a multi-event video from collapsing, we calculate the average cosine similarity. If the text features of one video are too similar, then the average cosine similarity among texts is expected to be closer to $1$. 
We find that in Me-Retriever(avg), the cosine similarity averaged on the whole test set is $0.789$ with a variance of $0.0059$ while in CLIP4Clip(mean), the value is $0.864$ with a variance of $0.0028$. 
And as in Fig.~\ref{fig:compare}, Me-Retriever has lower average cosine similarity than CLIP4Clip.
This shows that text features of the same video generated by Me-Retriever are more diverse than those by CLIP4Clip, which agrees with our motivation.

\section{Limitations}
Constrained by time and labor expenses, our experimental studies on the MeVTR task are based on the pre-existing video-text datasets including ActivityNet Captions and Charades. Nonetheless, it is important to recognize the potential bias in these datasets due to their emphasis on human activities and limitations in representing a wider range of videos. We aim to create a more comprehensive MeVTR benchmark encompassing more general videos.

\section{Conclusion}
We study a practical Video-Text retrieval scenario, Multi-event Video-Text Retrieval, in this paper. We find previous models suffer notable performance degradation in inference on MeVTR and textual feature collapse when retrained on MeVTR.
We present a new CLIP-based model, Me-Retriever, to conquer this challenge. In our experimental studies, we demonstrate the efficacy of these techniques on both Text-to-Video and Video-to-Text tasks for ActivityNet Captions and Charades-Event datasets. We believe this can be a strong baseline for future studies.

\section*{Acknowledgements}
\noindent
This work has been funded by the German Federal Ministry of Education and Research and the Bavarian State Ministry for Science and the Arts. The authors of this work take full responsibility for its content.

{\small
\bibliographystyle{ieee_fullname}
\bibliography{references}
}

\end{document}